%% file: main.tex
\documentclass[conference]{IEEEtran}
\input{cypher-settings.tex}
\IEEEoverridecommandlockouts                              

\usepackage{graphicx} 
\usepackage{silence}  							
\usepackage{hyperref}
\usepackage[utf8]{inputenc}    
\usepackage[T1]{fontenc}       
\usepackage[style=ieee,doi=false,isbn=false,date=year,backend=biber,maxbibnames=15,maxcitenames=2,mincitenames=1,uniquelist=false,uniquename=false,giveninits=true]{biblatex}
\usepackage[nolist,nohyperlinks]{acronym}
\usepackage{subcaption}
\usepackage{graphicx}
\usepackage{booktabs}
\usepackage{amsmath}
\usepackage[capitalize]{cleveref}
\usepackage{dsfont}
\usepackage{color}
\usepackage{colortbl}
\usepackage{xcolor}
\usepackage{array}

\bibliography{references2}
\title{What Did I Learn? Operational Competence Assessment for AI-Based Trajectory Planners}
\author{Michiel Braat$^{1}$, Maren Buermann$^{1}$, Marijke van Weperen$^{1}$, Jan-Pieter Paardekooper$^{1,2}$ 
\thanks{$^{1}$Netherlands Organisation for Applied Scientific Research, Integrated Vehicle Safety Group, 5700 AT Helmond, The Netherlands.}
\thanks{$^{2}$Radboud University, Donders Institute for Brain, Cognition and Behaviour, P.O.Box 9010, 6500 GL, Nijmegen, The Netherlands}}
\date{January 2025}
\begin{document}
\maketitle

\input{Chapters/0_abstract}

\input{Chapters/1_introduction}

\input{Chapters/2_background}

\input{Chapters/3_knowledge_graph}

\input{Chapters/4_competence}

\input{Chapters/5_experiment_1}

\input{Chapters/6_experiment_2}

\input{Chapters/7_conclusion}

\printbibliography

\end{document}

%% file: cypher-settings.tex
\usepackage{listings}
\usepackage{xcolor}

\lstdefinelanguage{Cypher}{
  morekeywords={MATCH, RETURN, WHERE, CREATE, DELETE, SET, MERGE, ON, OPTIONAL, WITH, LIMIT, SKIP, ORDER, BY, ASC, DESC},
  sensitive=true,
  morecomment=[l]{//},
  morecomment=[s]{/*}{*/},
  morestring=[b]",
  morestring=[b]'
}

%% file: Chapters/0_abstract.tex
\begin{abstract}
Automated driving functions increasingly rely on machine learning for tasks like perception and trajectory planning, requiring large, relevant datasets. The performance of these algorithms depends on how closely the training data matches the task. To ensure reliable functioning, it is crucial to know what is included in the dataset to assess the trained model's operational risk.\\
We aim to enhance the safe use of machine learning in automated driving by developing a method to recognize situations that an automated vehicle has not been sufficiently trained on. This method also improves explainability by describing the dataset at a human-understandable level.\\
We propose modeling driving data as knowledge graphs, representing driving scenes with entities and their relationships. These graphs are queried for specific sub-scene configurations to check their occurrence in the dataset. We estimate a vehicle's competence in a driving scene by considering the coverage and complexity of sub-scene configurations in the training set. Higher complexity scenes require greater coverage for high competence.\\
We apply this method to the NuPlan dataset, modeling it with knowledge graphs and analyzing the coverage of specific driving scenes. This approach helps monitor the competence of machine learning models trained on the dataset, which is essential for trustworthy AI to be deployed in automated driving.

\end{abstract}

%% file: Chapters/1_introduction.tex
\section{Introduction}
Automated driving (AD) functions are increasingly becoming integral to the modern transportation system. At the same time, these AD functions are incorporating more and more machine learning algorithms, such as deep neural networks (DNNs), for tasks ranging from perception to trajectory planning. These algorithms are often black box algorithms, making it challenging to understand their internal workings and predict their behavior in all contexts.

To achieve public acceptance of AD functions that include machine learning algorithms, it is important that they are trustworthy. A system is trustworthy when stakeholders are justified in putting trust in the system \cite{kastner_relation_2021}. For example, when a passenger trusts the automated lane keeping system functionality of their vehicle, this trust should be justified. Whether a stakeholder is justified to put trust in a system depends on factors such as whether the system is predictable, robust, aware of its own capabilities, and can explain its own actions or reasoning \cite{kastner_relation_2021, nordhoff_perceived_2021, khastgir_calibrating_2018}. However, trustworthiness does not mean that a trustworthy system always knows what to do, but rather that the trust placed in the system should align with the capabilities of the system \cite{nordhoff_perceived_2021, khastgir_calibrating_2018}. A trustworthy system should be able to identify when it is and when it is not competent. It must be possible to determine whether the system's correct functionality can be guaranteed in the specific context it is operating in.

The performance of machine learning algorithms depends heavily on how close a task is to what the algorithm was exposed to during training \cite{torralba_unbiased_2011}. Therefore, for trustworthy and reliable functioning of these algorithms, it is key to know what is included in the dataset and what is not, such that we know what can be expected from the trained model and determine its operational risk. Ensuring the safe deployment of machine learning functions involves recognizing when a vehicle encounters a situation that it has not been adequately trained for. This recognition is essential for preventing potential failures and enhancing the overall trustworthiness of AD systems. 

It is crucial for machine learning algorithms that the training data is of good quality and diverse. Therefore, increasing explainability and insight on what is actually in the driving data that a DNN is trained is essential. This shows the need of being able to describe the driving data at a human-understandable level, and making it easier to see exactly what a model has been trained and tested on.

We aim to accommodate the safe use of machine learning functions in AD by developing a competence measure that can estimate if an automated vehicle is entering a situation that it has not been sufficiently trained for, and should not be relied upon. This competence measure combines exposure to similar situations during training, or in other words the coverage of the current scene in the training dataset, and the complexity of the situation. This approach assumes that more complex scenes take longer to learn compared to simpler ones, necessitating greater coverage in such instances.

To make possible to estimate the competence, an accurate description of the situation the vehicle is in should be present. We propose to model driving data as knowledge graphs (KGs), depicting driving scenes at given time instances. Ultimately, by describing the data that is in the training set on a symbolic level in the KGs, and using it to calculate the complexity and coverage for scenes in the dataset, it is possible to create more insight in what is in the driving data.

In this paper, we demonstrate how we construct this competence measure based on learning experience through the following steps. \cref{section_knowledgegraph} details how the KGs are constructed, after which \cref{section_competence} explains how the competence metric is calculated based on the graphs. In \cref{section_dataset_insight}, we illustrate that we can compute the coverage and complexity of the scenes in the NuPlan Dataset \cite{caesar_nuplan_2022} to create more insight into what is in the dataset, and in \cref{section_competence_on_model}, we show that the competence metric for scenes correlates with the performance of a trained trajectory planner on the dataset.

%% file: Chapters/2_background.tex
\section{Background}\label{section_background}
\subsection{Competence Monitoring}
Competence monitoring involves assessing the reliability and accuracy of a model's predictions in real-time. Paardekooper et al. \cite{paardekooper_hybrid-ai_nodate} address competence estimation in a hybrid situational awareness system based on the level of doubt present in the current situation.

Closely related to confidence prediction is the measurement of model uncertainty. When a reliable uncertainty measure is available, it can be used to estimate the confidence of the model. Several methods exist to estimate the uncertainty of a DNNs predictions, such as Monte Carlo Dropout \cite{gal_dropout_2016}, Deep Ensembles \cite{lakshminarayanan_simple_2017} and Bayesian deep learning \cite{daxberger_bayesian_nodate}. However, a disadvantage of these techniques is that they require modifications to the model itself, either during training or execution. These approaches also significantly increase the computational time required for model inference.

Out-of-distribution (OOD) detection aims to determine whether a data sample falls outside the distribution of the training set. OOD detection is closely linked to competence estimation, as DNNs tend to under perform when a sample is not within the trained distribution. Ren et al. \cite{ren_likelihood_2019} propose a likelihood ratio method to improve OOD detection by correcting for background statistics that confound the detection process.

\subsection{Graph modeling}\label{subsec:Graph}
Graphs are suitable for modeling road networks, due to their ability to model the relationships between entities in such a way that makes the application of algorithms for further analysis meaningful and efficient \cite{yu_scene-graph_2022}.

Knowledge graphs (KGs) are graphs with additional semantics added to them. They contain types, labels, and properties to further enrich the meaning of nodes and the edges between them within the graph structure. KGs offer a structured way to represent various data sources and to enrich them with pre-established knowledge \cite{htun_integrating_2024} in a way that allows further computational processing, while also being human readable.

Neo4j is an open-source graph database management system designed to store, manage, and query highly connected data efficiently. Unlike traditional relational databases, Neo4j uses a schemaless property graph model, where data is represented as nodes, relationships, and properties. This structure enables fast traversal and querying of complex relationships, making it ideal for applications like road networks \cite{hristoski1_graph_2021}. Neo4j uses Cypher, a declarative graph query language optimized for graph operations, and provides scalability, flexibility, and can be used for real-time data analysis \cite{hristoski1_graph_2021, noauthor_neo4j_nodate}.

\subsection{Dataset Coverage}\label{subsec:B_Coverage} 
In \cite{piziali_functional_2004}, Piziali describes coverage as a "measure of verification completeness". This concept is used in different fields of research and engineering to measure how much of the defined requirements have been met. Examples for this include the amount of code that has been formally verified against expected properties (formal verification) \cite{gupta_formal_1992}, the number of driving scenarios or road conditions have been checked to ensure the reliability of an automated vehicle (AV verification) \cite{koopman_challenges_2016} or what percentage of the training data has been validated for quality and representativeness (AI model verification) \cite{liang_advances_2022}.

De Gelder et al. \cite{de_gelder_coverage_2024} introduces two types of coverage metrics as a measure of verification completeness for the purpose of AV verification. Their Type I coverage metric tries to measure how much of the operational design domain has been covered by a set of recorded driving scenarios, while the Type II metric is meant to capture how much of the driving data is covered by the catalog of defined scenario categories.\\
In an approach similar to their Type I coverage, this metric could be used to measure how many instances of a road context, modeled as a KG, have been covered within the data used for training a route planner, as proposed in this paper.

\subsection{Scene Complexity}
Only investigating the coverage of certain contexts within the training set would assume that all contexts are created equally, as if an AV (and even a human driver) can learn to navigate all types of contexts with the same ease. This assumption ignores the fact that both machine learning algorithms and humans require more practice to handle more complex situations. A more realistic approach is to include estimates about the complexity of a given driving scene.

Liu et al. \cite{liu_complexity_2024} propose several ways of quantifying complexity of driving scenes. Their approach combines both complexity estimates of static (e.g. road layout) and dynamic road (e.g. other vehicles) elements. In their paper, the authors discuss two types of complexity metrics. A \emph{static scene complexity quantification} and a \emph{dynamic scenario complexity quantification}. Since in our methodology, we are modeling contexts as static snapshots within a KG, we have chosen to adapt the static scene complexity quantification into our assessment. 


%% file: Chapters/3_knowledge_graph.tex
\section{Knowledge Graph Scene Construction}\label{section_knowledgegraph}
 The representation of the driving context in driving data typically consists of both a representation of the road infrastructure and the other actors and objects which surround the ego vehicle. The representation is mostly in a geometrically represented form on a 2-d plane \cite{shi_motion_nodate, li_scenarionet_nodate, gao_vectornet_2020} where either a grid map is used, or a vector representation using points, line-strings and polygons. The model has to interpret the relations between the different actors and pieces of the infrastructure itself from this view. 
 
 To formalize the driving context, we will use the term \emph{scene}. A \emph{scene} is a representation of the objects present in the surroundings of the ego vehicle and the most important relations between them. For this, a KG representation is used. Entities are represented in a graph using nodes. Each node is equipped with a collection of labels and attributes to distinguish it from other nodes within the graph. Relationships between nodes are modeled using edges, with each edge connecting a pair of vertices. Like nodes, edges are also distinguished using labels and attributes. Labels are used to identify types of nodes/edges, whereas attributes are used to distinguish between nodes and edges that share a label.

\begin{figure}
	\centering
	\includegraphics[width=\linewidth]{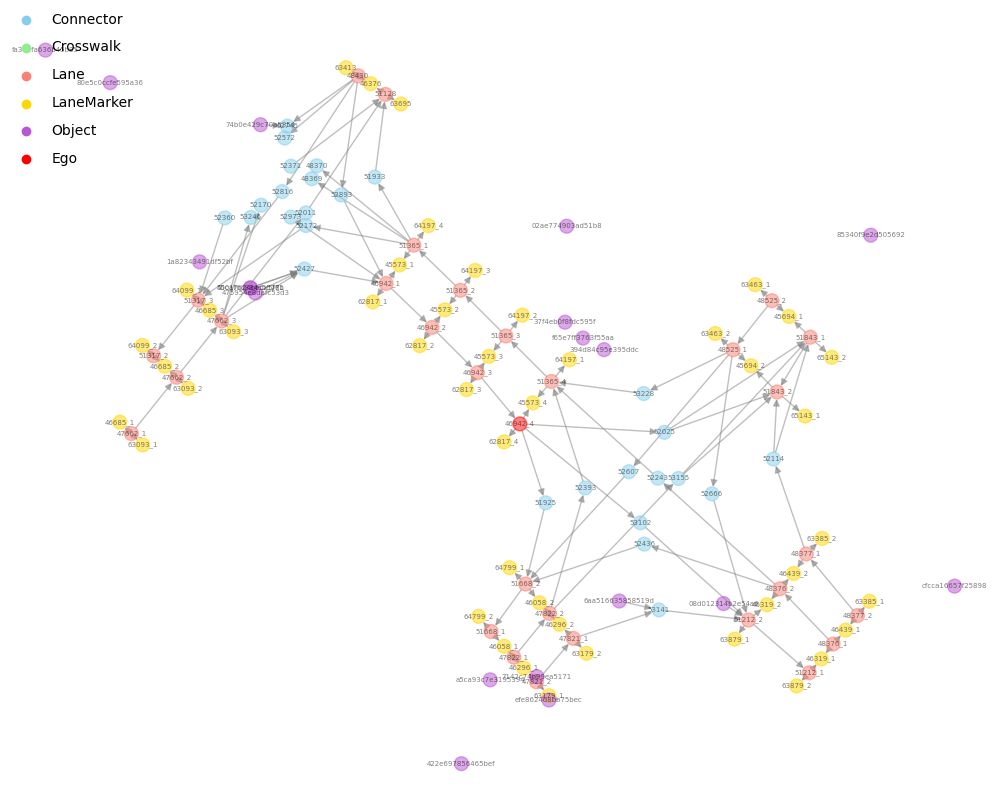}
	\caption{Example of a scene in the KG representation from the NuPlan dataset.}
	\label{fig:kg_example}
\end{figure}

The KGs representing the scenes are constructed by taking a snapshot of the world model data. First, the surrounding map is parsed. Infrastructure elements whose center points are within 50 meters of the ego vehicle are taken into account. The representation of roads and crossings is done based on lane segments and connections between lanes, similar to Lanelet2 \cite{poggenhans_lanelet2_2018} and the NuPlan representation. To standardize the representation of the lanes in the graph, lanes are segmented into parts with a maximum length of 10 meters. This segmentation is done in such a way that each lane on a road has the same number of segments and their endpoints are connected. Lane lines are segments in a similarly and connect adjacent lane segments.  All intersections are modeled using connector nodes, which connect lanes to each other that are accessible according to the intersection layout. Lanes are connected to neighboring lanes via lane lines to the left or right hand side of the lane. Other infrastructure elements include and are currently limited to crosswalks, but this can be extended to include elements such as traffic lights, stop lines and traffic signs etc. These elements are connected to lanes or connector nodes, to indicate their location within the road layout.

When the infrastructure is parsed, the ego vehicle and other actors are added to the graph. This is done by checking which lane segments had the most overlap with their bounding box. Information about the kinematics and type of actor are added as node attributes. \cref{table:kg_nodes} and \cref{table:kg_edges} lists all the different types of nodes, edges, and attributes that can occur in a scene graph. An example of such a scene from the NuPlan dataset \cite{caesar_nuplan_2022} can be seen in \cref{fig:kg_example}.

\begin{table}[]
\centering
\caption{Node types for scene KGs.}
\label{table:kg_nodes}
\begin{tabular}{ll}
\toprule
\textbf{Node Type}        & \textbf{Attributes}                               \\ \toprule
lane        & id, speed limit, length                  \\ 
connector   & id, turn type, length                    \\ 
lane marker & id, boundary type                        \\ 
crosswalk   & id                                       \\ 
ego         & id, velocity, dimension                  \\ 
object      & id, type, distance, velocity, dimensions \\ \bottomrule
\end{tabular}
\end{table}

\begin{table}[!h]
\caption{Edge types for scene KGs.}
\label{table:kg_edges}
\begin{tabular}{llll}
\toprule
\textbf{Edge Type} & \textbf{From Node Types} & \textbf{To Node Types} & \textbf{Attributes} \\ \toprule 
next                   & lane, connector          & lane, connector        &                     \\ 
connected to           & lane, connector          & lane marker            & side                \\ 
on                     & crosswalk, ego, object   & lane, connector        &                     \\ \bottomrule
\end{tabular}
\end{table}

For the experiments explained in \cref{section_dataset_insight} and \cref{section_competence_on_model}, a part of the NuPlan dataset is processed in this manner. The NuPlan dataset includes parsed scenes from Singapore and Boston, captured at 0.5-second intervals and stored in a Neo4j database. Each timestamped scene is represented by a KG. This process yields 72,480 scenes for Boston and 179,144 scenes for Singapore.

%% file: Chapters/4_competence.tex
\section{Competence}\label{section_competence}

The competence metric we calculate is meant to assess whether the system’s correct functionality can be guaranteed
in the specific context it is operating in. In order to do so, we estimate if a driving scene occurred sufficiently often within the training set (coverage). This coverage then gets weighted by the estimated complexity of the context the scene has been extracted from, to account for the fact that more complex scenes may require a higher coverage. A high competence score can be achieved when the coverage is high, and the complexity of the scene is low. For a more complex scene, a higher coverage is therefore needed and for less complex scenes, a lower coverage will suffice for achieving adequate competence results. That is, the competence of an AV-algorithm for a driving scene \emph{s} can be computed using:
\begin{equation}\label{eq:competence}
    \mathrm{Competence}(s) = \mathrm{Coverage}(s) \cdot (1-\mathrm{Complexity}(s))
\end{equation}
\begin{figure*}[!h]
	\centering
	\includegraphics[width=.8\linewidth]{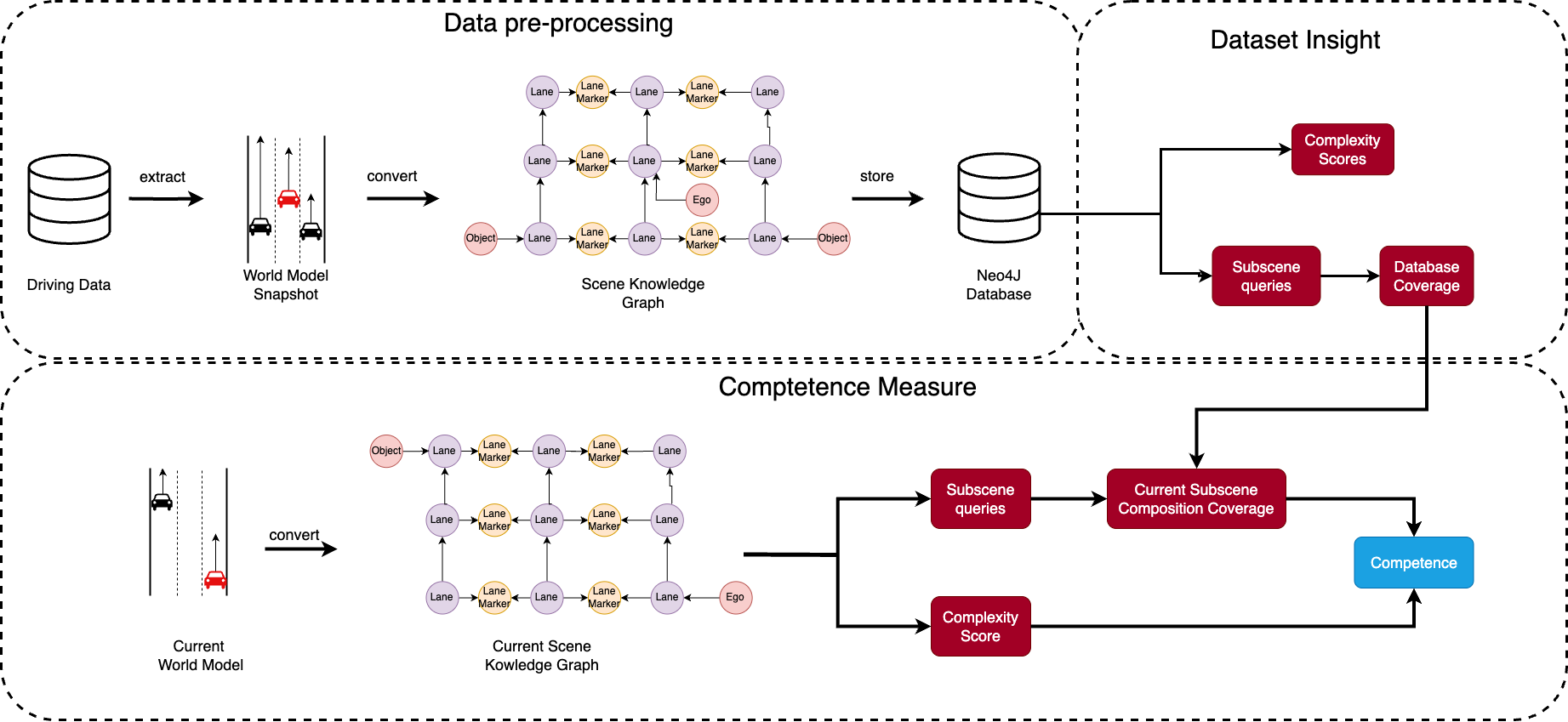}
	\caption{Flowchart illustrating the process of data pre-processing and competence measurement for driving data.}
	\label{fig:methododlogy}
\end{figure*}

\subsection{Matching graph sub-scenes}\label{subsec:sub-scenes}
The entirety of the graph represents the scene in which the ego is finding itself in. In this graph we search for specific patterns and count those to calculate the coverage. These patterns are called sub-scenes and represent situations within the scene that the ego vehicle needs to be able to handle. Sub-scenes can be extracted from the KG using Cypher queries.  For example, if we want to extract ``straight road'' sub-scenes from the KG, we can search for two consecutive lane nodes like this:
\begin{lstlisting}
MATCH (source:Lane)-[]->(target:Lane)
SET source:SubsceneNode, target:SubsceneNode
\end{lstlisting}
The \emph{MATCH} clause states that we are searching for a Lane node that has an outgoing edge to another Lane node. The \emph{SET} clause then marks all nodes that match this pattern. This approach of marking nodes allows to concatenate multiple queries together to match nodes that belong to more complex patterns (e.g., 4-way intersections and T-intersections). After marking the nodes of a sub-scene, we count the number of sub-scenes matching the queried pattern within the database in which the ego vehicle is currently present. That is, the node where the root attribute is set to True.
\begin{lstlisting}
    MATCH (n:SubsceneNode {root:True}) RETURN count(n) as result
\end{lstlisting}
In total, we query for 9 separate sub-scene patterns:
\begin{itemize}
    \item ego vehicle driving on straight road
    \item ego vehicle driving on roundabout
    \item ego vehicle enters roundabout
    \item ego vehicle leaves roundabout
    \item ego vehicle crosses intersection
    \item ego vehicle approaches intersection
    \item ego vehicle approaches pedestrian crossing
    \item vehicle driving ahead of ego vehicle
    \item vehicle driving behind ego vehicle
\end{itemize}
This is the minimal set of sub-scenes that we assumed to be relevant based on what data is currently modeled in the KG. These sub-scenes are not mutually exclusive, e.g., ``vehicle driving ahead of ego vehicle'' can occur with all other sub-scenes. Therefore, we also consider the composite occurrences of these sub-scenes, to describe the scene the ego vehicle is currently navigating as accurately as possible. This way, we can build a set of scenes $S$, where each $s \in S$ contains at least one sub-scene pattern. Scenes for which no sub-scene is found we label as unknown.

\subsection{Dataset Sub-scene Coverage}
The coverage metric used for estimating the driving scene competence is adapted from the \emph{Type I coverage metric} proposed in \cite{de_gelder_coverage_2024}. The function $\mathrm{Coverage}(s)$ maps a scene $s \in S$ to its corresponding coverage result. The coverage is computed using \cref{eq:coverage}, where $c(s)$ returns the number of times that a scene with the same set of sub-scenes as $s$ has been found within the training set and $ n \in \mathds{Z}^+$ is a hyperparameter indicating the minimum desired number of occurrences needed for a context to be considered to have sufficient coverage:
\begin{equation} \label{eq:coverage}
    Coverage(s) = \frac{\min (n, c(s))}{n}.
\end{equation}

\subsection{Complexity Assessment}\label{sec:ComplexityAssessment}
In this work, the complexity calculation is adapted from the \emph{static scene complexity quantification} proposed in \cite{liu_complexity_2024}. Here, the complexity of driving scenes is defined by three key components: natural environmental conditions, road conditions, and dynamic entities in driving scenes. The complexities of these three components are denoted as $c_1$, $c_2$, and $c_3$, respectively. The final \emph{static scene complexity score} can then be expressed using:
\begin{equation}\label{eq:complexity}
    Complexity(s) = \frac{c_1+c_2+c_3}{3}.
\end{equation}
In this work, all three components are adapted from \cite{liu_complexity_2024} in such a way that they can be compute using the data that is available in the KG. Additionally, the components are scaled between [0,1] by using a Min-Max normalization based on the complexities found in the whole dataset according to equation \cref{eq:minmax}, where $c'_i$ is the original complexity component value for every scene $i$ and $C'_i$ is the set of all $c'_i$ values in the dataset. 

\begin{equation}\label{eq:minmax}
c_i = \frac{c'_i - \min(C'_i)}{\max(C'_i) - \min(C'_i)} 
\end{equation}

\subsubsection{Environment complexity $c_1$}
We approximate the complexity of the environment through the number of unique elements present within the scene. This is expressed through the number of nodes with unique labels, as well as the number of unique node attributes that are indicative of the type of the element. For instance, a node with the label ``Object'' contains a type attribute, which could be another vehicle, pedestrian, traffic cone, etc.\\
Let us define \( V \) a the set of all nodes in the graph, where each node \( v \in V \) has: 

\begin{itemize}
    \item A \textbf{label} \( \lambda(v) \), where \( \lambda: V \to L \), and \( L \) is the set of all labels.
    \item  A \textbf{type attribute} \( \tau(v) \), where \( \tau: V \to T \), and \( T \) is the set of all type attributes.
\end{itemize}

Then, the total number of unique elements $U$, from both labels and type attributes in a scene, is given by:
\begin{equation}\label{eq:environment_complexity}
c'_1 = \left| \{ \lambda(v) \mid v \in V \} \cup \{ \tau(v) \mid v \in V \} \right|
\end{equation}

Using \cref{eq:environment_complexity}, we can then compute the environment complexity $c_1$, by applying a min-max normalization as done in \cref{eq:minmax}.

\subsubsection{Road obstacle complexity $c_2$}
While the environment complexity considers the number of unique types of elements within the scene, the road obstacle complexity considers the number of distinct objects in the scene that explicitly describe the presence of a road obstacle.\\
Let \( V_{\mathrm{O}} \subseteq V \) be the subset of nodes with the label \texttt{"Object"}, such that: 
\[V_{\mathrm{O}} = \{ v \in V \mid \lambda(v) = \text{"Object"} \}\]
 Let \( T_a \) be the set of object types, that describe road obstacles:
    \[
    T_a = \{\text{"generic\_object"}, \text{"traffic\_cone"}, \text{"barrier"}\}
    \]
Then, the subset of nodes $O \subset V_{\mathrm{O}}$, that describe road obstacles can be expressed as:
\[O = \{ v \in V_{\mathrm{O}} \mid \tau(v) \in T_a \}\]
Thus, the final expression for determining the distinct number of road obstacles becomes:

\begin{equation}\label{eq:obstacle_complexity}
c'_2 = \left| O \right|
\end{equation}

The final road obstacle complexity $c_2$ is computed by applying the min-max normalization from \cref{eq:minmax} on $c'_2$.
\\
\subsubsection{Dynamic entities complexity $c_3$}
The complexity score for dynamic entities tries to capture the level of attention a human driver would need to pay to its surroundings based on the number, speed, and distance to other road users within the scene. For estimating the complexity score, we consider the current velocity of the ego vehicle, $V_{\mathrm{ego}}$, since a higher velocity elevates the need for fast reaction times, and therefore increases the complexity of interacting with other road users.

In addition to  $V_{\mathrm{ego}}$, the longitudinal ($x$) and lateral ($z$) distance to other traffic participants ($m$) in the scene increase the complexity, as shorter distances reduce the time available to adequately react to sudden changes in target trajectories. For each traffic participant, we also differentiate between the types of participants, since each type of road user comes with its own set of maneuvers and challenges. The complexity values for each type of traffic participant ($x^{\mathrm{type}}$) are constant values and adapted from \cite{liu_complexity_2024}.

Given the estimates for $x_{\mathrm{type}}$, we can then compute $c'_3$ as:
\begin{equation}\label{eq:dynamic_entities}
    c'_3(s) = V_{\mathrm{ego}} \cdot \sum_{j=1}^{m} (0.5e^{-|x_j|}+0.5e^{-|z_j|}) \cdot \mathrm{ln}(1+e^{x^{\mathrm{type}}_j}),
\end{equation}
where \cref{eq:dynamic_entities} is adapted from Liu et al. \cite{liu_complexity_2024}. Using \cref{eq:dynamic_entities}, we can then compute the dynamic entities complexity $c_3$, by applying a min-max normalization as done in \cref{eq:minmax}.

%% file: Chapters/5_experiment_1.tex
\section{Results - Nuplan Dataset insight}\label{section_dataset_insight}
In this section we are analyzing two of the NuPlan data sets for the city of Boston and Singapore, which each contain around 120 h worth of driving data (more information about the data can be found in \cite{caesar_nuplan_2022}).

\subsection{Coverage}\label{subsec:coverage}
\begin{figure}[!h]
	\centering
	\includegraphics[width=\linewidth]{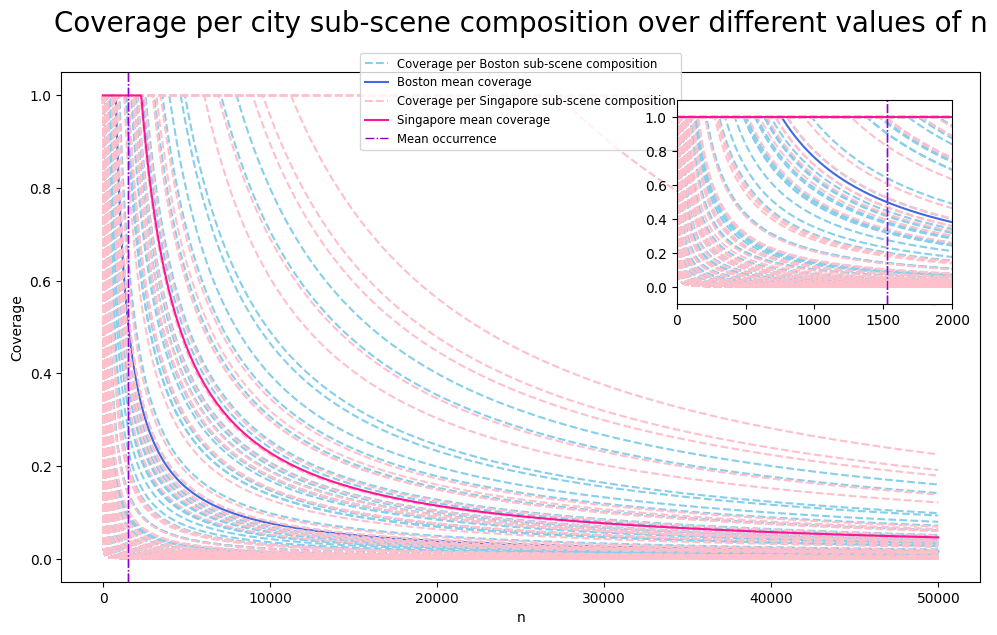}
	\caption{Coverage per city sub-scene composition over different values of n: The x-axis represents different values of $n$ (ranging from 0 to 50,000), while the y-axis shows the coverage (ranging from 0.0 to 1.0). The graph illustrates how coverage decreases with increasing $n$ for sub-scenes in Boston and Singapore, including their mean coverage and overall occurrences.}
	\label{fig:Coverage}
\end{figure}

After applying the sub-scene queries from \cref{subsec:sub-scenes}, the coverage for each (composite) sub-scene can be estimated. While \cref{tab:bostonVsingapore} shows how often each individual sub-scene pattern was found within the data, \cref{fig:Coverage} shows the mean coverage of all composite sub-scenes within the data of each city respectively for different hyperparameter values of $n$. The selection for n determines the threshold at which the number of sub-scenes is considered to be ``sufficient''. To keep the resulting coverage values comparable between the sub-scenes within and across the two data sets, the mean of all composite sub-scene counts has been chosen as a fixed value for n, which captures a wide range of coverage values between zero and one, as shown in \cref{fig:Coverage}. Additionally, \cref{fig:Coverage} shows the spread of the coverages per composite sub-scene. That is, the combination of the sub-scene elements from \cref{tab:bostonVsingapore} that occurred within the data.

Overall, the Singapore dataset contains more contexts, which results in a higher coverage for many of the (composite) sub-scenes, compared to the Boston data. On average, each composite sub-scene within the Boston set occurs less than 1000 times, while they occur on average more than 2000 times in the Singapore data.


\begin{table}[!h]
\caption{Number of sub-scenes that are occurring by themselves or within composite sub-scenes within the Boston and Singapore data.}
\centering
\renewcommand{\arraystretch}{.9} 
\begin{tabular}{@{}lll@{}}
\toprule
\textbf{Sub-Scene Element} & \textbf{Singapore} & \textbf{Boston} \\  \toprule
\multicolumn{1}{l}{straight road} & \multicolumn{1}{l}{116838} & \multicolumn{1}{l}{50826} \\ 
\multicolumn{1}{l}{approach intersection} & \multicolumn{1}{l}{108818} & \multicolumn{1}{l}{47256} \\ 
\multicolumn{1}{l}{on intersection} & \multicolumn{1}{l}{114956} & \multicolumn{1}{l}{44614} \\ 
\multicolumn{1}{l}{vehicle driving behind} & \multicolumn{1}{l}{32714} & \multicolumn{1}{l}{32568} \\ 
\multicolumn{1}{l}{vehicle driving ahead} & \multicolumn{1}{l}{19539} & \multicolumn{1}{l}{24571} \\ 
\multicolumn{1}{l}{approach crossing} & \multicolumn{1}{l}{5149} & \multicolumn{1}{l}{11173} \\ 
\multicolumn{1}{l}{enter roundabout} & \multicolumn{1}{l}{130} & \multicolumn{1}{l}{104} \\ 
\multicolumn{1}{l}{leave roundabout} & \multicolumn{1}{l}{91} & \multicolumn{1}{l}{65} \\ 
\multicolumn{1}{l}{on roundabout} & \multicolumn{1}{l}{287} & \multicolumn{1}{l}{50} \\ 
\end{tabular}%

\label{tab:bostonVsingapore}
\end{table}

While roundabout structures are the rarest in either city, in Singapore there are a total of 508 composite contexts, in which the ego vehicle either enters, leaves or navigates through a roundabout, compared to the 219 occurrences within Boston. However, while the Boston data contains less (composite) sub-scenes overall, it does contain more instances of vehicles driving ahead of the ego vehicle, as well as more composite sub-scenes in which the ego vehicle is approaching a crossing.

\begin{figure}[!h]
	\centering
	\includegraphics[width=3cm]{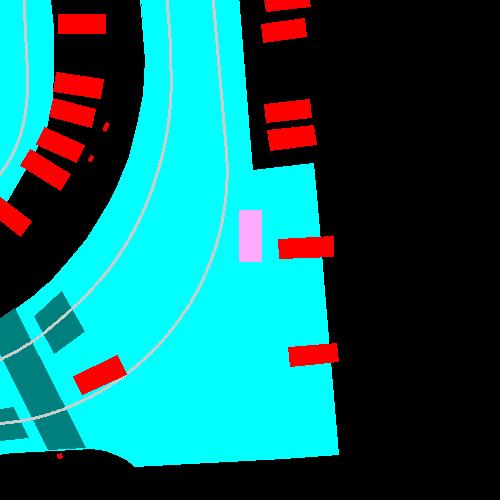}
	\caption{Example of a context with no sub-scene matches.}
	\label{fig:Unknown}
\end{figure}

Lastly, it should be noted that there is a substantial number of ``Unknown'' sub-scenes in both the Boston (1618 instances) and Singapore (5457 instances) data. This happens when the context modeled within a KG does not return any matches to any of the sub-scene queries. An example of such a case can be seen in \cref{fig:Unknown}, as the ego vehicle appears to be on a parking lot, rather than any of the other infrastructure elements for which we query.

\subsection{Complexity}\label{subsec:complexity}
\begin{figure}[!h]
	\centering
	\includegraphics[width=\linewidth]{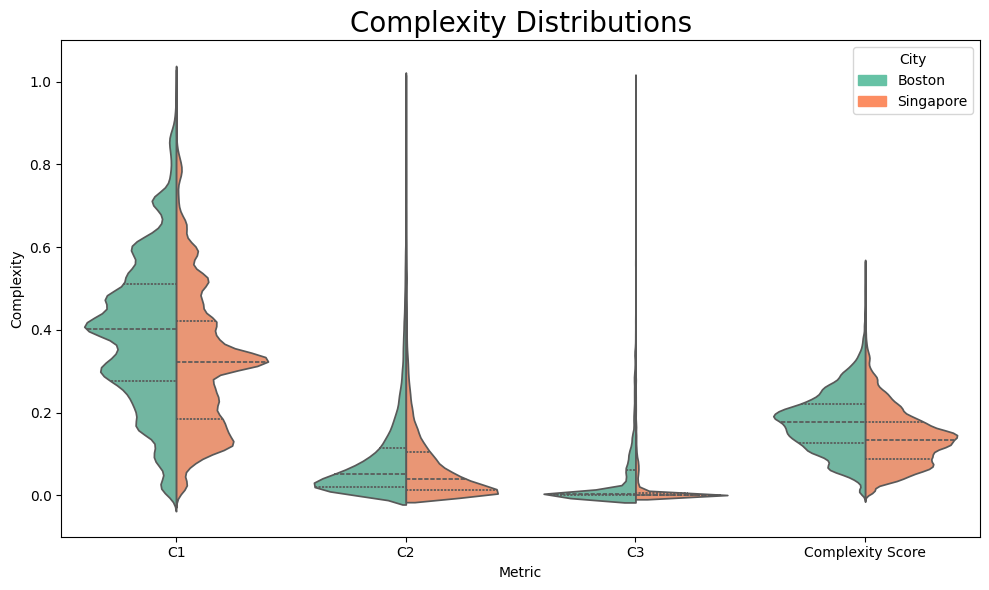}
	\caption{Violin plots illustrating the complexity distributions across three metrics ($C_1$, $C_2$, and $C_3$) and the total complexity for Boston and Singapore.}
	\label{fig:Complexity}
\end{figure}

After computing the values for $C_1$, $C_2$ and $C_3$, the final complexity scores can be computed, as discussed in \cref{sec:ComplexityAssessment}. \cref{fig:Complexity} shows the distributions of complexity scores for each complexity component for both Boston and Singapore. While the distributions of each of the complexity components span from zero to one, due to the normalization that has been applied, the final complexity score reaches a maximum value of ca. 0.55.

For each of the complexity metrics, as well as their average (illustrated by the final complexity score), it can be noted that the distributions of complexity score are rather similar between the two cities.

For all complexity values, the contexts present in the Boston data achieve slightly higher complexities than the contexts in the Singapore data. The composite sub-scene containing the sub-scenes: `intersection', `vehicle driving behind', `vehicle driving ahead', `approach intersection', `straight road', illustrates why this is. For both Singapore and Boston, this composite sub-scene returns the largest complexity values. However, within the Boston data, contexts with this composite sub-scene occur 2544, whereas they only occur 1518 times in the Singapore data. Since the part of the Boston map selected for data collection in the NuPlan dataset is more urban than its Singaporean counterpart, it seems logical that this would be the case.

%% file: Chapters/6_experiment_2.tex
\section{Results - Competence measure of trajectory planner}\label{section_competence_on_model}
In this section, we demonstrate how the competence measure can be used to assess the competence of a trajectory planner. To accomplish this, a DNN trajectory planner with the Autobot \cite{girgis_latent_2022} architecture was trained on Singapore data from the NuPlan dataset. The Boston data from the NuPlan dataset was used for the evaluation of the competence metric and the comparison to the performance of the model. This means that the dataset used in the coverage calculation of the competence metric is only based the Singapore part of the NuPlan dataset.

\begin{figure}[!h]
	\centering
	\includegraphics[width=\linewidth]{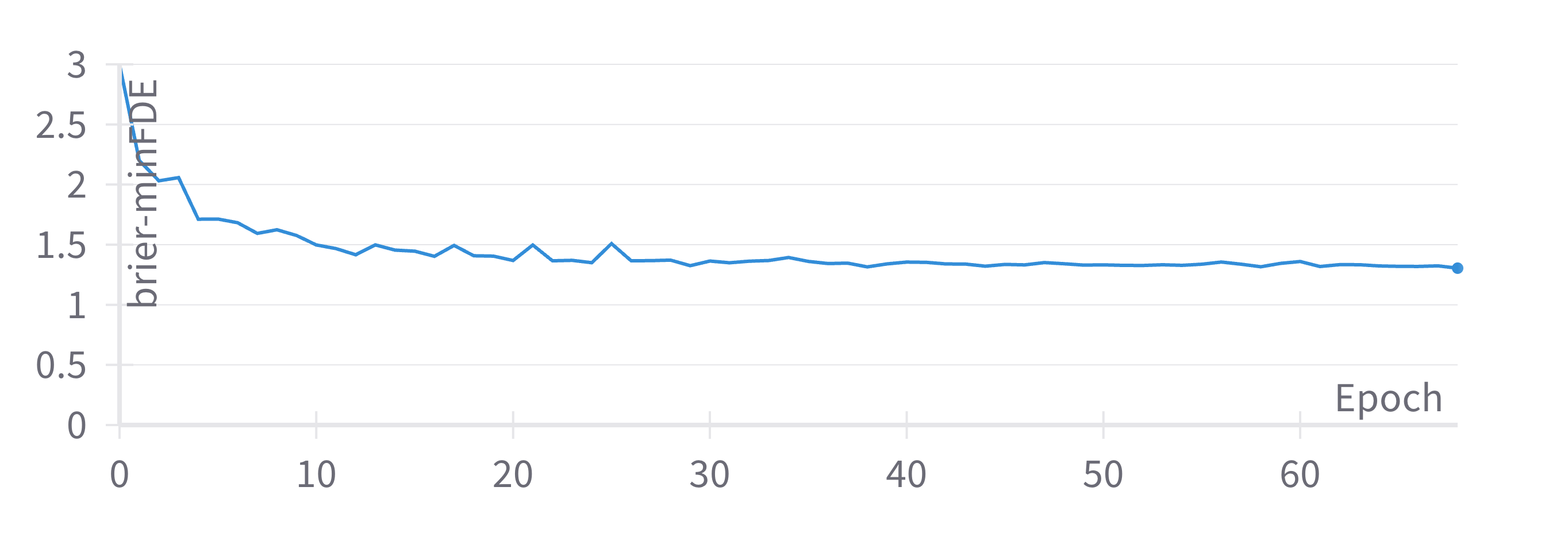}
	\caption{Validation loss of the Autobot trajectory planner.}
	\label{fig:autobot_training_validation}
\end{figure}

The training of the Autobot trajectory planner was done using the UniTraj \cite{feng_unitraj_2024} framework. After 68 epochs, the validation loss did not decrease any further, and early stopping was used. \cref{fig:autobot_training_validation} shows the validation loss development during training.  

The following four common evaluation metrics for trajectory prediction were used. 

\begin{itemize}
    \item The Miss Rate is defined as the ratio of trajectories for which the minFDE is greater than two meters. 
    \item The Minimum Average Displacement Error (minADE) is the mean Euclidean distance between the points of the ground truth trajectory and the predicted trajectory.
    \item The Minimum Final Displacement Error (minFDE) is similar to the minADE, but only the final point in the trajectory is taken into account.
    \item The Brier Minimum Final Displacement Error (brier-minFDE) is the minFDE but it takes into account the prediction confidence of the trajectory based on the Brier score.
\end{itemize}

\begin{figure}[!h]
	\centering
	\includegraphics[width=\linewidth]{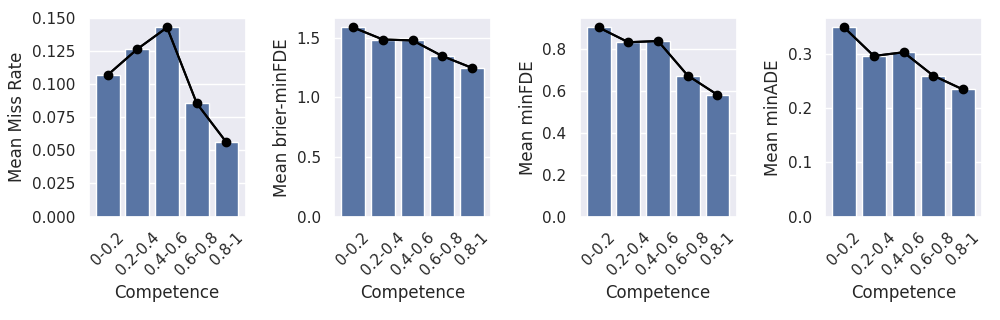}
	\caption{Mean value of the evaluation metrics over the competence metric binned in steps of 0.2.}
	\label{fig:autobot_bar}
\end{figure}

\begin{figure}[htbp]
    \centering
    \begin{subfigure}[b]{0.45\linewidth}
        \centering
        \includegraphics[width=\textwidth]{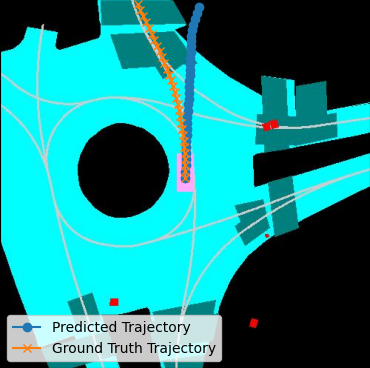}
        \caption{Competence: 0.00}
        \label{fig:low_comp_low_performance}
    \end{subfigure}
    \hspace{0.05\linewidth}
    \begin{subfigure}{0.45\linewidth}
        \centering
        \includegraphics[width=\textwidth]{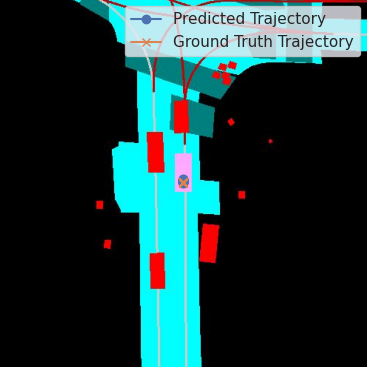}
        \caption{Competence: 0.10}
        \label{fig:low_comp_good_performance}
    \end{subfigure}
    \begin{subfigure}[b]{0.45\linewidth}
        \centering
        \includegraphics[width=\textwidth]{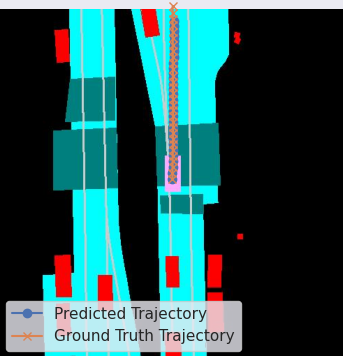}
        \caption{Competence: 0.84}
        \label{fig:high_comp_good_performance1}
    \end{subfigure}
    \hspace{0.05\linewidth}
    \begin{subfigure}[b]{0.45\linewidth}
        \centering
        \includegraphics[width=\textwidth]{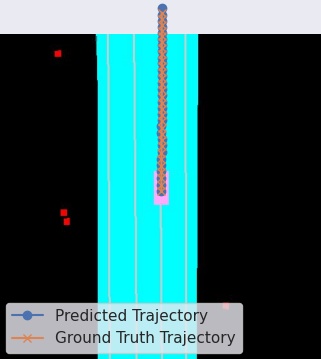}
        \caption{Competence: 0.95}
        \label{fig:high_comp_good_performance2}
    \end{subfigure}
    \caption{Examples of low and high competence scenes with trajectories.}
    \label{fig:main}
\end{figure}

Additionally, the competence metric was calculated for these scenes. The Pearson correlation was calculated between the competence and these evaluation metrics. We assume there is a relationship between them because the trajectories should be accurate when the competence is high, but unpredictable when the competence is low. \cref{table:autobot_correlation} shows that there is a weak but significant negative correlation between the evaluation metric and the competence metric. \cref{fig:autobot_bar} shows the mean value of the evaluation metrics over the competence and \cref{fig:main} shows four trajectory examples with the calculated competence. As expected, most of the values decrease as the competence increases, except for the mean Miss Rate, which increases for the lower competence bins of [0-0.2] to bin [0.4-0.6]. 

These results show that overall there is a relation between the competence metric and the actual performance of the trajectory planner. The fact that it is only a weak correlation and that the Miss Rate increases for lower level competencies can be caused by multiple factors. First, low competence means that the situation is unseen and possibly complex. This implies that the output of the model is unpredictable and should not be trusted. However, it can still be the case that the model infers a correct trajectory. This can be seen in \cref{fig:low_comp_good_performance}, where the competence is low, but the model is still able to correctly predict a trajectory that remains stationary behind the vehicle ahead. Second, parts of the context that influence the performance in these cases might not be modeled in the KG or in the sub-scenes. To alleviate this problem, more context should be modeled and additional important sub-scenes can be added.

\begin{table}[]
\centering
\caption{Pearson correlation between evaluation metrics and competence metric for the Boston data.}
\label{table:autobot_correlation}
\begin{tabular}{@{}lll@{}}
\toprule
\textbf{}                          & \textbf{correlation}        & \textbf{p-value}            \\ \toprule
\multicolumn{1}{l}{Miss Rate}    & \multicolumn{1}{l}{-0.103} & \multicolumn{1}{l}{0.0012} \\               
\multicolumn{1}{l}{brier-minFDE} & \multicolumn{1}{l}{-0.090} & \multicolumn{1}{l}{0.0046} \\ 
\multicolumn{1}{l}{minFDE}       & \multicolumn{1}{l}{-0.087} & \multicolumn{1}{l}{0.0059} \\ 
\multicolumn{1}{l}{minADE}       & \multicolumn{1}{l}{-0.092} & \multicolumn{1}{l}{0.0035} \\ \bottomrule
\end{tabular}
\end{table}

%% file: Chapters/7_conclusion.tex
\section{Conclusion}\label{section_conclusion}
In this work, a novel competence metric has been developed for assessing machine learning functionality based on the context that the automated vehicle is situated in. This metric consist of two different aspects, the coverage of the scene in the training dataset, and the complexity of the scene. Both are based on a symbolic representation of the context using a KG. Experimental results show that the competence metric aligns with the performance of a trained trajectory planner. Additionally, we have shown that both the coverage and complexity aspects can be used to describe and compare what is in a driving dataset, providing valuable insight into the dataset. The approach to competence measurement described in this work is a valuable tool to enable safe AI by predicting whether the trajectory planning output is trustworthy.

\subsection{Future work}
A weakness of the current competence metric is that according to \cref{eq:competence}, the competence is always 0 when the complexity of the scene is 1, even if there is a high coverage for this scene. This was not experienced in the experiments done in this paper, as can be seen in Figure \ref{fig:Complexity}, but \cref{eq:competence} should be adapted that a high coverage can still lead to a high competence, even in scenes with a complexity of 1. 

On top of this, we aim to build upon our current method to further improve the KG model and extend the relations and objects it encompasses. This could help to cover and describe more of the context, resulting in a better competence metric. Additionally, the list of sub-scene queries can be extended to cover more sub-scenes, as the current list of sub-scenes that is queried for is limited. Expanding the number of sub-scenes would reduce the number of scenes categorized as unknown in the coverage and increase the accuracy of the coverage metric by enabling more sub-scene configurations to be queried.

Furthermore, the current model of the  context uses static snapshots of the environment. This ignores the temporal interactions and relations that are relevant to the driving task. Future work should incorporate the time dimension into the KG model, allowing for the modeling of entire driving scenarios rather than just static context snapshots.

\subsection{Acknowledgment}
This work was funded by the Horizon Europe programme of the European Union, under grant agreement 101076754 (project AITHENA). Views and opinions expressed here are however those of the author(s) only and do not necessarily reflect those of the European Union or CINEA. Neither the European Union nor the granting authority can be held responsible for them.